# The language and social behavior of innovators

Fronzetti Colladon, A., Toschi, L., Ughetto, E., & Greco, F.

# The Language and Social Behavior of Innovators

**Abstract**

Innovators are creative people who can conjure the ground-breaking ideas that represent the main engine of innovative organizations. Past research has extensively investigated who innovators are and how they behave in work-related activities. In this paper, we suggest that it is necessary to analyze how innovators behave in other contexts, such as in *informal* communication spaces, where knowledge is shared without formal structure, rules, and work obligations. Drawing on communication and network theory, we analyze about 38,000 posts available in the intranet forum of a large multinational company. From this, we explain how innovators differ from other employees in terms of social network behavior and language characteristics. Through text mining, we find that innovators write more, use a more complex language, introduce new concepts/ideas, and use positive but factual-based language. Understanding how innovators behave and communicate can support the decision-making processes of managers who want to foster innovation.

**Keywords:** Social Network Analysis; Emotional Text Mining; Innovation; Innovators; Organizational Communication.

# 1. Introduction

For decades, a big question that has driven innovation research is "How can managers find innovative people for their organization?". In the corporate context, such innovators are creative people who can conjure ground-breaking ideas that drive competitive advantage and business success. A wide stream of literature has sought to answer this question by scrutinizing how such individuals discover innovative ideas when trying to solve problems, and thereby define a pool of skills that characterize innovative people.

However, less attention has been given to this question: *How do innovators differ from non-innovators in their behavior when they are not engaged in searching for innovative solutions?* In addition to the behavioral traits that innovators show during the discovery phase of innovations, it is important to direct attention to other characteristics that they may show when acting freely in their work-related social interactions. The motivation for adopting this perspective is simple: the greater the number of employees showing innovator characteristics, the more opportunities there are for the organization to achieve innovation and high performance (George & Zhou, 2007; Park et al., 2014; Zhou & George, 2001). This investigation may help an organization better identify potential innovators among its employees by analyzing their way of interacting and communicating in informal contexts. The word "innovation" itself, which is a combination of "in" and "nova", emphasizes the concept that the generation of new ideas occurs *in* the organization (Y. Lee et al., 2018) and, thus, stresses the key role played by employees, who represent a strategic constituency of any organization (Kim & Rhee, 2011). Surprisingly, we are not aware of any extant research that has adopted this perspective so far.

For the past few decades, many empirical studies have focused on observing innovators' specific behaviors and personality features in order to understand how these traits differ from those of non-creative individuals (Dyer et al., 2011; Fürst & Grin, 2018; Kandemir & Kaufman, 2019). The psychological literature has found that innovative employees possess distinctive characteristics and information processing habits, such as associating seemingly unrelated problems, questioning in the right way to understand phenomena, observing like anthropologists and social scientists do, experimenting by creating prototypes, and networking with diverse experts to obtain radically different perspectives (Dyer et al., 2011). They also show a higher level of education, higher competencies, diverse work interests and duties, self-confidence, and a tolerance for ambiguity (Maddi, 1996). Moreover, social network theorists suggest that innovators show a higher centrality in advice networks (P. A. Gloor et al., 2020): The structure of their social relationships determines the quantity and quality of their knowledge, as well as how rapidly people can acquire and elaborate the information necessary to discover new opportunities (Garud & Prabhu, 2020). Finally, recent research has recognized the strategic importance of the voluntary communicative behavior of innovators: collecting relevant information and generating knowledge are two catalytic resources for increasing an organization's effectiveness and innovative performance (Kim & Rhee, 2011; Mazzei, 2010; Slotnick, 2007; Sofiyabadi et al., 2020; Ye et al., 2016). However, all these studies tend to focus on the behavioral traits that enable innovators to discover novel ideas.

The current paper seeks to complement this stream of research by adopting a new perspective: one focused on the role of *informal* spaces of communication, where knowledge is shared within the organization without any formal structure, rules, and work obligations. In this way, we hope to more broadly disentangle the traits behind innovator DNA. Specifically, we investigate how innovators behave when they do not search for innovation or directly try to solve problems, but rather when their spontaneous desire for engagement leads them to interact with other organizational employees to share any information – related or not to their work life.

To this end, we analyze the communicative behaviors of innovators in the intranet forum of a large multinational company[1], where a group of employees regularly and freely interact to share experiences, seek and provide advice, and share knowledge, ideas, and personal opinions. We analyzed about 38,000 forum posts written by more than 11,000 employees over 18 months. This setting allowed us to analyze how innovators (the 0.4% of our sample) interact with different organization members (i.e., in different areas and at different hierarchical levels) across multiple discussions. The forum unites a network of individuals characterized by heterogeneous ideas, perspectives, and motivations, who freely decide to participate in discussions and interact with the community without explicitly searching for innovation. By combining social network analysis, semantic analysis, and text mining techniques, we explain how innovators differ in terms of: (i) social network behavior as disclosed by their online forum interactions and (ii) language characteristics in terms of semantic metrics and content.

Our contribution to the literature is twofold: First, we build on and integrate disparate literature strands examining innovators' DNA, their role within organizations, and communicative style. While prior studies have separately described the main traits or behaviors that depict innovators, we cast new light on how language interacts with behavior. Second, we extend the knowledge on innovators' characteristics in new directions, beyond the formal setting of the organization they work for. In fact, by combining new and innovative metrics of text mining to analyze their language (e.g., emotional text mining), alongside social network analysis tools (e.g., *in-* and *out-distinctiveness* centrality) to explore their behaviors, we shed light on how the DNA of innovators unfolds within an informal space of organizational communication.

---

[1] We cannot disclose the name of the company for confidentiality reasons.

Our findings show that innovators write longer and more complex posts, featuring more elaborate diction and syntax. While their communication has a positive sentiment, it is rather objective and fact-based. Innovators introduce new words/topics in the discourse and are outward-oriented – in the sense that they look beyond the company boundaries and consider possibilities of growth and performance improvement. Non-innovators, on the other hand, maintain an inward look: Their communication is often oriented toward achieving personal goals in terms of relational or economic advantages. Innovators are also more active and dynamic in their social interactions. They write more posts and are more central in the social network. Moreover, their contacts are more distinctive, so they potentially provide knowledge to questions that would otherwise remain unanswered.

The paper is organized as follows: Section 2 discusses the theoretical background. Section 3 introduces our methodology and data. Section 4 presents the results and Section 5 concludes.

**2. Theoretical background**

The expression "Eureka!" captures the moment when great ideas are discovered. However, innovation is often generated beforehand from frequent communications with different stakeholders, which allow one to gather critical information for refining and advancing ideas. Fittingly, organizational social network research often emphasizes how innovation arises from individuals' personal interactions. However, this literature still lacks a comprehensive analysis that combines advanced semantic metrics and social network tools to jointly examine innovators' social behavior and language, especially in more informal settings, such as blogs or corporate online forums. To elaborate on innovators' behaviors within an organization, this Section summarizes the main findings of the literature (see also Table 1A in the Appendix).

### 2.1. The communication style of innovators

As a social factor that guides individual and group cognition (Ocasio et al., 2018), language has the power to shape how individuals present themselves (O'Connor, 2002) and, in turn, how others view them. Narratives are also widely believed to be an effective strategy in innovation communication (Weber & Grauer, 2019). Thus, individuals often adjust their language in response to group differences within an organization (Bazarova et al., 2013; Gil-Lopez et al., 2018). Starting from these premises, several studies have explored the connections between communication behaviors and innovation within organizations (Allen et al., 2016; Fosfuri & Tribó, 2008; Raz & Gloor, 2007; Sousa & Rocha, 2019).

Scholars interested in how innovation spreads within organizations have shown that innovators are keen to communicate with others and less worried about a variety of communication situations than their colleagues (Dyer et al., 2011; Ray et al., 1996). Successful innovators are more passionate about conversation and curious to learn from others. This enthusiasm is reflected in their frequency of interactions, as well as the time and effort dedicated to communicating with others (P. A. Gloor et al., 2020).

Scholars across different disciplines (e.g., business, human resources, and communication management) have recognized the importance of studying communication as one crucial dimension of employees' behavior that ultimately affects organizational performance (Harandi & Abdolvand, 2018; Johnston et al., 2019). When combined with leadership, effective communication drives organizational change (Alonderiene et al., 2020; Hsing-Er Lin & McDonough, 2011; Schaufeli, 2013) and encourages organizational trust. Leaders who display transparent and honest communication are trusted by their followers (Xu & Cooper Thomas, 2011) and can better exert their influence within the organization in a meaningful way. In the leadership literature, the study of communication skills has been extended to the issue of employees' communication environment. Van Dyne, Ang, and Botero (2003) and Morrison (2014) conceptualized voice and silence as a function of employees' willingness to express (or not) ideas, information, and opinions about work-related improvements. An inclusive communication environment has been found to make employees feel more valued, thereby positively affecting their engagement within an organization (Rees et al., 2013; Reissner & Pagan, 2013).

In recent years, a limited number of works have started to analyze the uniqueness and distinctiveness of the language used by innovators, drawing on prior evidence about such individuals' information processing habits (Dyer et al., 2011). In a recent study, Gloor and colleagues (2020) focused on the formal exchange of emails within an organization. By examining the email archives of the R&D employees and managers of a US multinational corporation, they found that innovators have a higher number of interactions (i.e., send and receive more messages), are more committed in conversations, and have more direct contacts in the network. To linguistically distinguish innovators from non-innovators, Greco et al. (2021) introduced the "Crovitz Innovator Identification Method": a tool for identifying how a person's language signals innovative thinking. In this paper, we expand prior research by questioning: *Do innovators differ from non-innovators in the use of language?*

**2.2. The social behavior of innovators**

In recent years, the organizational and innovation literature has shown growing interest in the networks and relations between individuals, groups, and organizations. In tandem, the bulk of organizational social network research has adopted the network approach to study the social interactions of individuals within organizations – see the literature reviews by Carpenter et al. (2012), Dobrow et al. (2012), Phelps et al. (2012) and Muller and Peres (2019).

Organizational social network research has largely emphasized the study of the relations between actors embedded in a network (Cross & Prusak, 2002). The actors' roles within the network are shaped by a variety of relationships (both formal and informal) that link them and constitute the basis for social interactions. These relationships, in general, act as critical conduits for knowledge search, creation, and transfer within organization boundaries (Shane & Cable, 2002), enhancing information flow and reducing information asymmetries in a corporate context. Scholars have observed that individuals' position within the network is highly responsible for the rapid diffusion of information, as well as enhances both exploratory and exploitative innovation (Ibarra et al., 1993; Yan et al., 2020). Typically, individuals who occupy a central position in the network (i.e., having a large number of ties) have better information access and are more equipped to leverage this information to influence other network members (Cullen et al., 2018; P. M. Lee et al., 2011). Moreover, there is a strong nexus between network centrality and source of power (Tsai, 2001). Especially in innovation contexts, centrality plays a crucial role in the effective implementation and subsequent commercialization of research outputs and innovative work (Cangialosi et al., 2021; West & Richter, 2008). Having access to several sources of information is pivotal to innovation and is facilitated by the central positioning of individuals who act as innovators within an organization (Muller & Peres, 2019; Zasa et al., 2022). Central individuals can affect their peers' behavior either by withholding or modifying the knowledge to be shared with other network members (Ibarra et al., 1993), or by simply playing a certification role whereby they can observe others' actions, even without formally sharing that information (Deeds et al., 2004; Elfring & Hulsink, 2007; Gulati & Higgins, 2003; Shane & Cable, 2002). Central individuals are able to rapidly disperse their knowledge throughout the network, allowing it to become common knowledge for other members[2] (Valente, 1996).

One of the current advances in organizational social network research relates to the embodiment of individual attributes of actors into the understanding of the structuralist aspects of network foundations. In particular, the literature exploring the micro-foundations of organizational social networks has opposed a psychological versus sociological perspective by questioning whether individuals constitute the social networks they are embedded into or whether network patterns affect individuals' fate, shaping their characters and behaviors (Tasselli et al., 2015). The first approach views the underpinnings of individuals' network behaviors in their personality traits, cognition, and psychological characters. In this vein, network preferences and positioning depend upon each network participant's personality and individual differences (Battistoni & Fronzetti Colladon, 2014; Fang et al., 2015; Mehra et al., 2001). According to the second perspective, differences in individuals' behaviors and personalities are enabled and constrained by the structural patterns of the social network in which they are embedded (Landis, 2016). These arguments are particularly prominent in emerging innovation contexts. The co-evolutionary approach attempts to combine the two perspectives, postulating that individuals and the social networks they belong to reciprocally influence each other (Schulte et al., 2012). In other words, people's behaviors affect and are affected by their relationships within a network. At the same time, they contribute to shaping such relationships with their distinctive personalities, perceptions, and psychological traits (Kilduff et al., 2008).

---

[2] However, research has also suggested that having too many ties may be detrimental for an individual (Kwon & Adler, 2014), ultimately stifling their efficiency within an organization. Managing too many relationships, while informationally advantageous, can imply significant costs and resources (in terms of time and effort) (Elfring & Hulsink, 2007; Mariotti & Delbridge, 2012). After a certain threshold of managed relationships, the cost of maintaining them – coupled with information overload (Cooper et al., 1995)– overweighs the benefits deriving from an information advantage, through a defocusing of action and a reduction in the ability to alleviate information asymmetries (McFadyen & Cannella, 2004). Alternatively, a central network position might be disadvantageous because of individuals' feeling of needing to assist their ties, which might be time consuming and lead them to postpone other important accomplishments (McFadyen & Cannella, 2004).

Despite the growing number of studies focusing on social networks within organizations, there is no consensus among scholars about the "boundary conditions" that influence the benefits that individuals gain from network embeddedness. Indeed, some scholars have pointed out that the value of ties may vary depending upon the context (e.g., cultural/social context) (Gulati & Higgins, 2003). In addition, it remains unclear how the structural characteristics of the network (cohesion versus structural holes) facilitate or constrain the flow of resources and information within the network (Kilduff & Brass, 2010). While a cohesive network facilitates the formation of trust, information exchange, and the flow of non-redundant resources among network members, social constraints may arise that limit network members' flexibility (Bellavitis et al., 2017). Factors that drive the trade-off between the benefits and costs of membership in a cohesive network—rather than a network rich in structural holes—have not been adequately explored (Bellavitis et al., 2017; Shipilov & Li, 2008). At the same time, the mechanisms behind groupthink dynamics are not yet clear. Within a network, sources of similarity among individuals (i.e., homophily, same cultural background, etc.) or other idiosyncratic aspects (i.e., passion, attention to ethical values, etc.)

are likely to drive the interaction matching (Fronzetti Colladon et al., 2021; Nedkovski & Guerci, 2021) and influence people's behavior. To explore the factors that shape the relationship among individuals in a structured network, we need a carefully nuanced analysis of the contextual and institutional conditions that may affect the outcomes of this relationship. Indeed, the field needs to further investigate the process, transformation, empowerment, enactment, operation, and dynamics that drive the formation and evolution of networks, as well as how these dimensions reflect the relationships between individuals (Zhang et al., 2008). In this paper, we expand prior research by asking: *Do innovators differ from non-innovators in their social interactions?*

## 3. Methodology

To carry out this study, we analyzed the online communication of more than 11,000 employees working for a large multinational company, the name of which will be kept anonymous for confidentiality reasons. These employees communicated online via a public intranet forum (only accessible to them), where they could share news about their work, seek and provide advice, share knowledge, exchange opinions and ideas and comment on the posts of others. Employees were also free to share their day-to-day experiences or discuss other topics – such as special discounts or recreational activities reserved for employees. In addition, the forum was used by the Human Resources (HR) team for internal communication and to advertise important news about the company. We collected and analyzed about 38,000 messages (written in Italian) posted over one and a half years[3].

About half of the employees using the online forum (5,773) contributed with a single post; 2,115 employees posted two messages, 1,008 posted three messages, and the remaining 2,122 posted four or more messages. The average post length was 91 words, with an SD of 441.69 words. Only 1.4% of employees received a public comment/answer to their posts. This number is only a partial indicator of reactions. Indeed, as we clarify in the limitations, we have no data regarding post visualizations/reads/likes or private messages exchanged among employees.

---

[3] We excluded from the analysis messages sent by the HR and internal communication staff, as their posts were made on behalf of the company.

The identification of innovators among the employees was carried out by an internal team comprising six representatives of the HR department. In this preliminary selection phase, short-cut decision-making metrics were used to reduce the long list of employees into a manageable set of plausible innovators. In particular, the team pre-selected employees based on the results of their last two annual performance reviews (one review was considered for a small number of recently hired employees). Reviews were made by direct managers, but also considering the feedback received by other team members. Each review had an overall score and only employees with ratings in the upper quintile were considered. The sample was further reduced by looking at the specific section where employee innovativeness was commented on and evaluated by their direct superiors.

After this initial assessment, a committee comprising four senior HR managers adopted larger set of screening criteria in order to focus their time and effort on a limited list of people. In this phase, the committee assessed an employee's propensity to suggest and implement ideas for better practices, procedures, and products for the company. Raters were instructed to evaluate individuals based on their involvement in innovative projects and the novelty of their work. Raters evaluated each employee independently, with an initial agreement on 85.7% of the cases. Subsequently, the committee met to discuss discordant cases to find common agreement.

Out of 11,018 employees who posted on the online forum, only 49 were classified by the company as top innovators.

### 3.1. Analysis of Social Behaviors

The first step in our analysis was to build a social network representing forum interactions, i.e., a graph with N nodes (one for each employee) and M arcs (connecting employees who answered each other). Accordingly, we have an arc starting at the generic node A and terminating at node B if A answered at least one post of B. Arcs were weighted with the number of interactions. The forum was organized into threads such that each employee could open a new discussion thread or post in existing ones. They could answer each other's posts without the possibility of mentioning other users, as one would do, for example, on Twitter or Facebook.

In order to evaluate employees' online behaviors, we used the methods and tools of social network analysis (Wasserman & Faust, 1994) to calculate a set of ego-network metrics. Specifically, we wanted to look at the position of innovators in online communication networks, studying their centrality from multiple angles. The concept of network centrality is well-known in social network analysis, with new metrics continuously being developed (Fronzetti Colladon & Naldi, 2020). Among all the possible measures, we selected the three most popular – i.e., degree, betweenness, and closeness – as they have been used for decades to capture different aspects of social actors' positions (Freeman, 1979; Wasserman & Faust, 1994). In particular, with degree and weighted degree, we analyze the online communication of employees and their degree of interactivity, looking at how many messages they send and receive and how many interactions they have with different peers. Closeness helps us evaluate employees' embeddedness in the social networks and the speed at which they can reach their peers, measured in terms of degrees of separation. Betweenness measures employees' brokerage power, i.e., their ability to create bridging connections that link their peers, which sometimes reflects the ability to access valuable sources of knowledge by spanning team and organizational boundaries (Cross et al., 2002; Marrone, 2010). Our choice of looking at these metrics is aligned with past research studying the online behavior of innovators and top performers, which proved a connection between centrality and performance (Ahuja et al., 2003; P. Gloor, 2017; P. A. Gloor et al., 2020; Wen et al., 2020).

In addition to betweenness centrality, we consider a similar well-known metric (Burt's *constraint* measure) that looks at the existence of *structural holes* in employees' ego-networks, i.e., missing links between contacts in a person's network that provide mediation opportunities and related advantages (Burt, 1995). Burt (2004) showed that brokerage across structural holes provides social capital and leads to higher compensation and performance. People who provide bridging connections have a greater chance of accessing the non-redundant information and resources that facilitate good ideas (Burt, 2004; Cowan & Jonard, 2007). Accordingly, we maintain that this metric is an important addition to betweenness centrality that establishes a more comprehensive perspective on brokerage. We similarly consider distinctiveness centrality, which is a more recent measure that adds information over traditional metrics (Fronzetti Colladon & Naldi, 2020). Distinctiveness shows that the centrality of a social actor may depend on not only its social position, but also on that of its neighbors. It departs from the traditional idea (Hansen et al., 2020) that being connected to very popular nodes is more beneficial than having links with peripheral social actors. Distinctiveness looks at centrality from another angle and shows the importance of reaching loosely connected peers, as the distinctiveness of these connections can provide access to unique sources of knowledge.

A more detailed explanation of each metric and its calculation is provided in the following.

*In-degree*. This measure counts the number of incoming network ties of a node, i.e., how many different people answered the posts of a specific employee. In its weighted version, the weights of incoming arcs are summed. Therefore, *weighted in-degree* counts the number of answers received by a forum user, regardless if they were written by a single person or many – complementing the information provided by in-degree. *Out-degree* and *weighted out-degree* consider the same information albeit for outgoing arcs. Accordingly, out-degree represents the number of peers an employee wrote to, while weighted out-degree is the total number of answers, news, or comments they posted.

In- and out-degree centrality consider the number and strength of direct social connections, but neither account for their uniqueness nor the degree of connected peers. Past research has suggested looking at the *distinctiveness centrality* of nodes to understand whether a node's connections reach loosely connected peers or are just another link to overly connected others (Fronzetti Colladon & Naldi, 2020). Indeed, links to the network periphery are important as they allow one to reach nodes that would otherwise remain isolated. Using the formula suggested by Fronzetti Colladon and Naldi (2020), we calculated *in-* and *out-distinctiveness* centrality on our directed graph. In-distinctiveness centrality attributes higher importance to nodes that receive incoming connections from peers who send few outgoing arcs. In an online forum context, this corresponds to receiving answers from people who only answer to us (or to few other people). On the other hand, out-distinctiveness places higher value on the outgoing arcs that reach nodes with few incoming connections. In the online forum, this means valuing the role of employees who answer the posts/questions of colleagues who receive answers from few others (or from nobody else).

*Closeness*. Closeness centrality represents the extent to which a node can quickly reach its peers in the network. Distances are calculated by considering the shortest network paths between each possible pair of nodes; closeness is calculated as the reciprocal of the distance of a node from all others (Wasserman & Faust, 1994). The measure is normalized considering the maximum number of direct connections, N-1. A node with maximum closeness (score of 1) would be directly connected to all other nodes.

*Betweenness*. This measure counts how often a node lies between the shortest paths connecting all other nodes (Wasserman & Faust, 1994). Having high scores of betweenness centrality is usually indicative of high brokerage power.

Lastly, we also calculated the value of Burt's *constraint* (1995), representing the extent to which a social actor has a higher number of open or closed triads in its ego-network. Indeed, when an employee can mediate a connection between unlinked peers, we say s/he has a *structural hole* in its ego-network, which usually leads to mediation advantages and brokerage power. Structural holes can support faster career advancement, the combination of knowledge and generation of new ideas, or the implementation of divide-et-impera strategies (Burt, 2004). On the other hand, when employees cannot act as a mediator (because all their peers are directly connected), they are fully 'constrained' by the social structure of their connections. Accordingly, constraint scores vary from 0 to 1, with higher scores indicating smaller freedom and fewer structural holes.

### 3.2. Language characteristics

The study of language complements that of social behaviors (P. Gloor, 2017). Accordingly, we used well-known indicators of text mining and semantic analysis to describe the main characteristics of language used by each employee. Firstly, we calculated how much each employee wrote in total (counted as the total number of words across all their posts). We called this variable *Word Count*. Instead, the variable *WPS* measures the average number of words per sentence.

*Six-letters*. This variable measures the percentage of words that are longer than six letters. It has been used in past research to discriminate between simple and complex language, with the idea that texts with longer words indicate a higher language complexity (Tausczik & Pennebaker, 2010).

*Sentiment*. Sentiment reflects the positivity or negativity of the language used in forum posts. Scores vary from -1 to 1, where 1 is totally positive and -1 is totally negative; values around zero indicate neutral posts. In order to calculate this metric, we used the social network and semantic analysis software Condor (P. Gloor, 2017) and averaged the scores obtained for each post at the individual level.

*Novelty*. Language novelty measures how much employees use new words in their posts, deviating from the common dictionary (i.e., the words not commonly used by their peers). Employees have high language novelty if they write about something new or very specific (using new terms) and if new terms are not buried under many other common words. This indicator has been used in past research (e.g., P. Gloor, Fronzetti Colladon, Giacomelli, et al., 2017; Wen et al., 2020) in relation to different dependent variables; it has proven, for example, to be a proxy of work-engagement (P. Gloor, Fronzetti Colladon, Grippa, et al., 2017). We calculated novelty by penalizing the use of words that appeared in many forum posts, using the term frequency inverse document frequency (TF-IDF) information retrieval metric (Jurafsky & Martin, 2008). In particular, we calculated the TF-IDF scores of the words used by each employee, comparing them with those used by their peers – such that if everybody is using the word "thanks", this will have a score of zero as it is an extremely common term. Subsequently, we calculated the average of all TF-IDF scores at the individual level to see how much each employee was using words outside the common language, without burying them under very long messages. We used the following formula:

$$Novelty = \frac{1}{n} \sum_{w \in V} f_w \, log_{10} \frac{N}{n_w}$$

where N is the total number of employees, $n$ is the total number of words that appear in the posts of an employee and $V$ is their set; $f_w$ is the frequency of word $w$ and $n_w$ is the number of employees who use the word $w$. Before calculating novelty (or counting words), we preprocessed the texts by removing stop-words and word affixes (stemming) (Bird et al., 2009; Jivani, 2011).

### 3.3. Emotional Text Mining

The second step involved extending the analysis of the language used by employees—not just referring to semantic metrics, but digging deeper into how innovators communicate, the words they use, the topics covered, and their associations. To this end, we used a method known as Emotional Text Mining (ETM) (Greco & Polli, 2020b) to profile individuals according to their communication.

There is a close connection between word use and behavior, even if this relationship is indirect (Greco & Polli, 2020b). ETM considers the relationships among words, with the idea that word associations are tightly related to mental functioning that ultimately determines behavior (Freud, 1989). Accordingly, ETM can be used to understand people's attitudes, expectations, and opinions. This profiling technique has been applied in different fields (Cordella et al., 2018)—and particularly to the analysis of online interactions—in order to profile people or anticipate their behavior (Greco et al., 2017; Greco & La Rocca, 2020; Greco & Polli, 2020a, 2020c). In this work, we extend the analysis of communication styles and social behaviors through ETM in order to comprehensively profile innovators working in the target company and distinguish them from non-innovators.

ETM performs a sequence of synthesis procedures, from corpus preprocessing to term selection and multivariate analysis, which are presented in the following and further detailed in the work of Greco and Polli (2020b). Textual data were cleaned and preprocessed with the software T-Lab. Before proceeding with the analysis, we calculated three lexical indexes to evaluate if it was possible to statistically process the data (Giuliano & La Rocca, 2010): the total number of words (3,022,411); the ratio of the number of different words (types) to the total number of words (3.6%), and the ratio of the number words occurring only once (hapaxes) to the total number of different words (hapax percentage = 50.4%). We removed stop-words and lemmatized words using the T-Lab dictionary (Lancia, 2020), thus reducing the overall number of terms. Subsequently, we selected terms of medium rank frequency to perform a multivariate analysis (Bolasco, 1999). We built a term-document matrix and performed a cluster analysis through a bisecting k-means algorithm based on cosine similarity (Savaresi & Daniel, 2004; Steinbach et al., 2000). Optimal clustering was determined by considering the Calinski-Harabasz, the Davies-Bouldin and the intraclass correlation coefficient ($\rho$) indices (Greco & Polli, 2020b). The last step consists of a correspondence analysis on the term-cluster matrix (Lebart et al., 1998) and a chi-squared test, the latter of which is used to assess differences in the distribution of posts between innovators and non-innovators. Correspondence analysis is used to represent the discourse through different factors—in our case leading to a bidimensional factorial map with two axes (factors). Each term is univocally assigned to a factor – thus allowing a subsequent labeling and interpretation – based on its absolute contribution to the inertia explained by that factor (Boccia Artieri et al., 2021). In other words, the terms are assigned to the factor they contribute to the most.

In the last step, each cluster was manually labeled by three text mining experts. They met and jointly examined the content of each cluster to find an appropriate label based on the cluster's most characterizing terms (Greco & Polli, 2020b). Labels are used to convey each cluster's meaning and main theme. The same procedure was applied to generate the labeling of the two axes of the factorial map (see Figure 1).

## 4. Results

### 4.1. Profiling innovators through Emotional Text Mining

The results of the cluster analysis show that the 944 keywords allow the classification of 91.5% of messages. The clustering validation measures show that the optimal solution is three clusters (Calinski-Harabasz = 239.4; Davies-Bouldin = 17.3; $\rho = 0.019$). In Table 1, we listed the ten most representative words for each cluster (identified through the chi-squared metric, which is used to determine if the assignment of each word to a cluster is statically significant).

| Cluster | Percentage Messages in the Cluster | Lemma | Translation | Chi-squared** |
|---|---|---|---|---|
| 1 Market and Innovation | 52.7% | europeo | European | 441.5 |
| | | milioni | millions | 394.6 |
| | | innovazione | innovation | 359.3 |
| | | *aspetto tecnico** | *word related to technical aspects** | 345.8 |
| | | business | business | 338.1 |
| | | paese | country | 326.6 |
| | | imprese | companies | 313.7 |
| | | infrastruttura | infrastructure | 307.1 |
| | | cittadini | citizens | 290.0 |
| | | miliardo | billion | 289.8 |
| 2 Social and Emotional Connection | 25.3% | ottimo | great | 11505.9 |
| | | complimenti | congratulations | 9635.9 |
| | | ciao | hi | 7219.7 |
| | | bravo | good | 4561.2 |
| | | sperare | to hope | 4327.6 |
| | | piacere | pleasure | 2780.4 |
| | | interessante | interesting | 1822.5 |
| | | notizia | news | 1713.4 |
| | | collega | colleague | 1578.1 |
| | | conoscere | meet | 1469.5 |
| 3 Corporate Benefits and Initiatives | 22.0% | e-mail | e-mail | 3817.6 |
| | | codice | code | 2862.8 |
| | | biglietto | ticket | 1780.0 |
| | | richiesta | request | 1668.2 |
| | | attivare | activate | 1664.3 |
| | | ricevere | receive | 1493.7 |
| | | acquisto | purchase | 1298.9 |
| | | inserire | to insert | 1298.1 |
| | | riuscire | to succeed | 1296.0 |
| | | link | link | 1253.2 |

\* The words revealing information about participants or the company were changed
(while keeping their meaning) in compliance with privacy agreements.
\*\* p < 0.001 for all the chi-squared tests reported in the table.
**Table 1**. Cluster analysis results

We can observe three types of communication: Employees who communicate on the topics of cluster 1 are mostly oriented to innovation and the market. They use words concerning business performance, innovation, and marketing, reflecting their concern for the company's growth and new business opportunities. Half of the messages (52.7%) belong to this cluster. The second and third clusters make up the other half of the corpus (47.3%). Employees communicating on the topics of cluster 2 care about social relationships: They interact to nurture their social connections within the company and build new informal links. For example, they show appreciation for other people's posts, thanking them or congratulating them on their achievements. Their posts add little knowledge or information to the discourse. People in the third cluster are mostly concerned with employees' benefits and little with company growth. For example, they care about the online activation of employee-dedicated company offerings or to the redemption of free tickets to attend company-sponsored events.

Most of the innovators' messages are innovation- and market-oriented (96%). They belong to cluster 1. Only 4% of their messages belong to the other clusters. Non-innovators, on the other hand, are much more focused on improving their social relationships within the company (27% of messages) and on employee benefits (23% of messages), with only 50% of messages related to market and innovation. A chi-squared test confirmed the significance of innovators and non-innovators' different distributions across topics ($\chi^2 = 1156.4$, df=2, p = 0.000).

In Figure 1, the three clusters are represented in a two-dimensional factorial space through correspondence analysis (Lebart et al., 1998). The numbers on the vertical and horizontal axes are the spatial coordinates, which help us understand if a cluster is distant or close to another. Similarly, deviation from the origin represents a cluster distance from the average discourse. Each graph axis represents one of the two factors, with the first (horizontal) explaining 66.7% of total inertia. Considering the words that are associated with the extremes of the two factors, one can label the two axes as "view" and "personal gain" (see Table 2 and Table 3). Employees have an inward "view" when their communication focuses on company events and an outward view when they pay attention to the world outside the company. The second factor – which we call "personal gain" and corresponds to the vertical axis – reflects a communication meant to achieve either a relational or personal advantage. As we can see from the figure, cluster 1 (that of innovators) is horizontally distant from the other two, thus making factor 1 the main distinction element between innovators and non-innovators. Innovators are not particularly interested in personal gain and company benefits; they are outward-oriented, i.e., look outside the company boundaries. Their conversations have a broad scope and are related to company growth and marketing opportunities. Non-innovators, on the other hand, seem to look at a smaller world, the one constrained by the company boundaries; they care about their personal benefits without considering that these are tightly related to the company's growth.

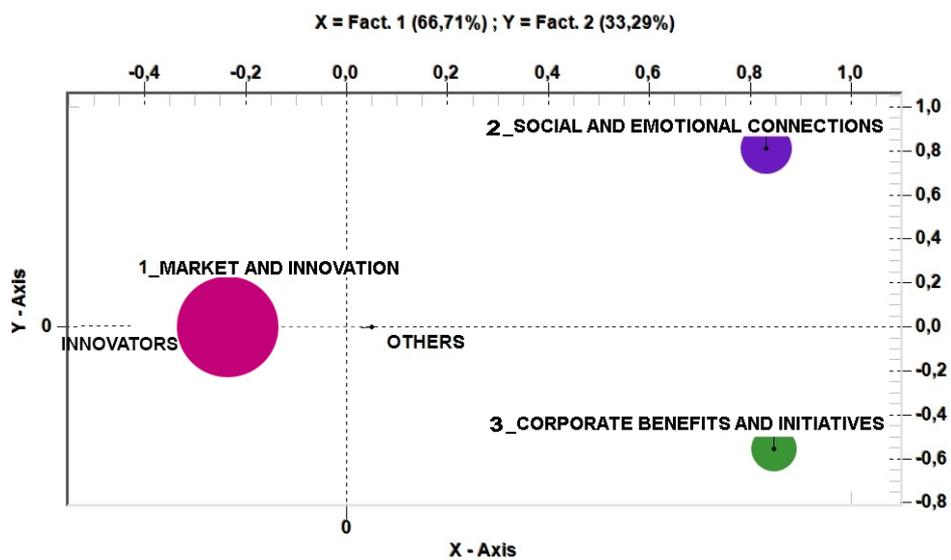

**Figure 1.** Factorial map

| Cluster | Percent Messages | Label | Factor 1 Positioning | Factor 2 Personal Gain |
|---|---|---|---|---|
| 1 | 52.7 | Market and Innovation | Outward | |
| 2 | 25.3 | Social and Emotional Connection | Inward | Relational |
| 3 | 22.0 | Corporate Benefits and Initiatives | Inward | Economic |

**Table 2**. Correspondence analysis interpretation

| Factor 1 (Horizontal Axis) | | | | Factor 2 (Vertical Axis) | | | |
|---|---|---|---|---|---|---|---|
| Outward View keyword | AC | Inward View keyword | AC | Personal Advantage keyword | AC | Relational Advantage keyword | AC |
| Innovation term* | 0.08% | E-mail | 2.20% | Purchase | 0.83% | Great | 11.42% |
| Big | 0.12% | Employee benefit* | 1.89% | Access | 0.62% | Congratulations | 9.55% |
| Infrastructure | 0.23% | Good morning | 1.56% | Buy | 0.61% | Hi | 5.92% |
| Development | 0.07% | Promotional code | 1.49% | Number | 0.60% | Bravo | 4.47% |
| Public administration | 0.08% | Obtain | 1.26% | To request | 0.53% | Hope | 3.81% |
| Regional | 0.11% | Activate | 1.07% | To apply | 0.40% | Nice to meet you | 2.85% |
| Billion | 0.22% | Colleague | 0.99% | Send | 0.38% | Interesting | 2.08% |
| European | 0.33% | Request | 0.94% | Credit card | 0.34% | Meet | 1.56% |
| Agenda | 0.15% | Ticket | 0.91% | Click | 0.33% | News | 1.55% |
| Companies | 0.23% | Function | 0.85% | Book | 0.29% | Greetings | 1.41% |
| Large | 0.17% | Promo name* | 0.85% | Confirmation | 0.28% | Relational term* | 1.32% |
| Citizens | 0.22% | Receive | 0.83% | Price | 0.28% | Speak | 1.08% |
| Grant | 0.06% | Technical term about benefit request* | 0.83% | Reply | 0.27% | Idea | 0.83% |
| City | 0.08% | Problem | 0.78% | Economic term* | 0.26% | Dear | 0.83% |
| Global | 0.08% | Application | 0.78% | Money | 0.25% | Pity | 0.73% |
| Decree | 0.15% | Link | 0.77% | Payment | 0.25% | Relational term* | 0.72% |
| Banda | 0.26% | Discount | 0.73% | Product | 0.24% | Remember | 0.69% |
| Institution | 0.07% | Promo | 0.72% | Address | 0.23% | Card | 0.56% |
| Adoption | 0.07% | Promo name* | 0.71% | Ask for economic benefit* | 0.23% | Heart | 0.52% |
| Healthcare | 0.07% | iphone | 0.71% | Through | 0.21% | Useful | 0.49% |

Note. AC is the absolute contribution of each term to the inertia explained by each factor.
* In compliance with privacy agreements, the words revealing information about participants or the company were changed (while keeping their meaning).

**Table 3**. Words characterizing the two factors of Figure 1

## 4.2. Profiling innovators through social network and semantic analysis

In our case study, we could profile innovators as a panel of experts had previously identified them. However, there might be cases where innovators are unknown and thus difficult to profile. For this reason, we extended our analysis and used the metrics of text mining and social network analysis presented in Section 3 to see whether innovators could be identified based on their distinctive communication style and/or social behavior.

We started with an exploratory analysis through t-tests, comparing the mean scores of innovators and non-innovators (see Table 4). We used Welch's t-tests to account for unequal group variances. We also calculated the significance of Mann-Whitney U tests to complement the analysis with a non-parametric approach.

| Variable | Group | Mean | Welch's t-test p-value | U Mann Whitney Significance |
|---|---|---|---|---|
| In-degree | Innovator | 26.63 | 0.064 | 0.000 |
|  | Non-innovator | 1.80 |  |  |
| Out-degree | Innovator | 5.65 | 0.000 | 0.000 |
|  | Non-innovator | 2.00 |  |  |
| Weighted in-degree | Innovator | 33.94 | 0.081 | 0.000 |
|  | Non-innovator | 2.83 |  |  |
| Weighted out-degree | Innovator | 25.29 | 0.005 | 0.000 |
|  | Non-innovator | 3.34 |  |  |
| In-distinctiveness | Innovator | 93.87 | 0.070 | 0.000 |
|  | Non-innovator | 6.57 |  |  |
| Out-distinctiveness | Innovator | 14.42 | 0.000 | 0.000 |
|  | Non-innovator | 2.69 |  |  |
| Closeness | Innovator | 0.72 | 0.000 | 0.000 |
|  | Non-innovator | 0.37 |  |  |
| Betweenness | Innovator | 8819.70 | 0.031 | 0.000 |
|  | Non-innovator | 259.50 |  |  |
| Constraint | Innovator | 0.43 | 0.000 | 0.000 |
|  | Non-innovator | 0.77 |  |  |
| Word Count | Innovator | 7719.53 | 0.056 | 0.000 |
|  | Non-innovator | 283.52 |  |  |
| Sentiment | Innovator | 0.13 | 0.000 | 0.654 |
|  | Non-innovator | 0.17 |  |  |
| Novelty | Innovator | 4.22 | 0.000 | 0.000 |
|  | Non-innovator | 1.71 |  |  |
| WPS | Innovator | 44.29 | 0.000 | 0.000 |
|  | Non-innovator | 18.78 |  |  |
| Six-letters | Innovator | 35.36 | 0.000 | 0.000 |
|  | Non-innovator | 29.51 |  |  |

*Note*. Independent-sample t-tests, with equal variances not assumed.

**Table 4**. Innovators vs. others, t-tests

As the table indicates, innovators are, on average, significantly different from other employees – for all variables except for word count, in-degree, weighted in-degree, in-distinctiveness, and sentiment (if we take a significance threshold of 5% and only consider results that are significant for both the Welch's and the Mann-Whitney's tests).

Innovators have a higher number of social interactions than non-innovators: They send a higher volume of messages and are more central within the network. Figure 2 shows the distribution of the number of posts for the innovators and non-innovators. The number of posts are more evenly distributed for innovators than for non-innovators. Employees writing more than 20 posts are the most frequent group among innovators, whereas about 50% of non-innovators wrote a single post.

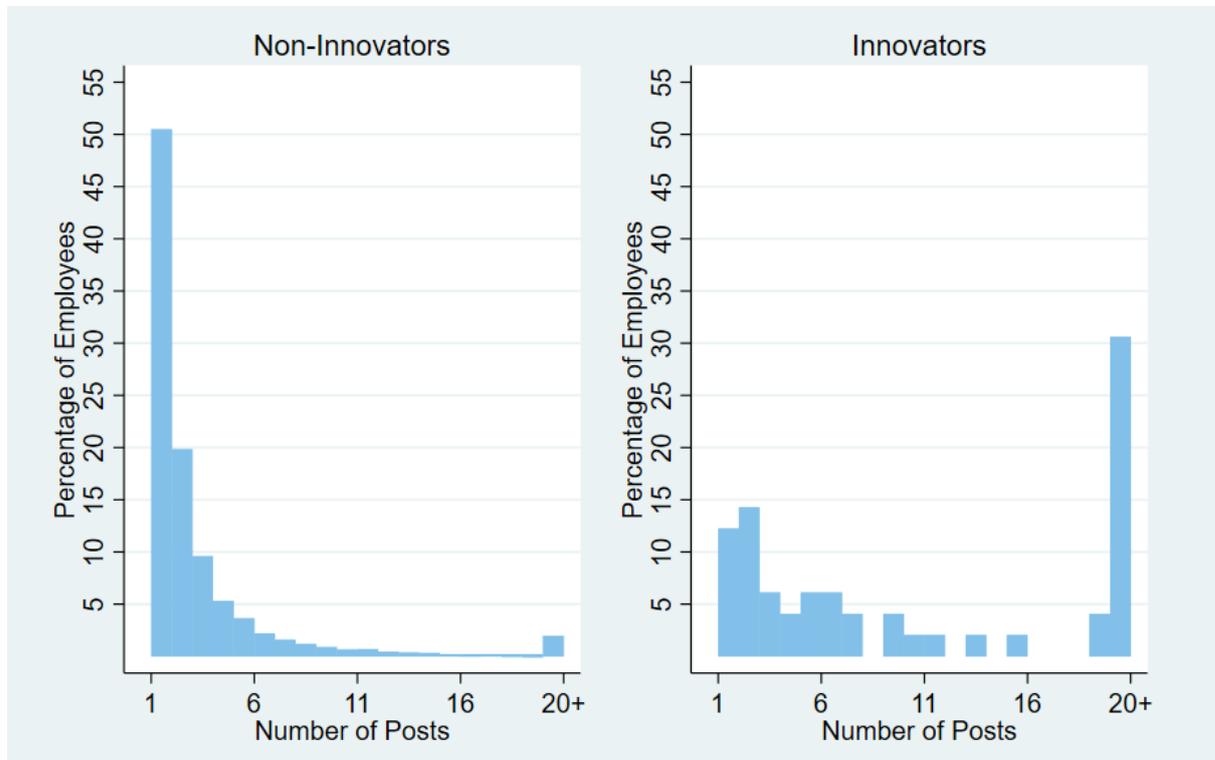

**Figure 2**. Distribution of the number of posts

Innovators also have more distinctive social connections, reaching nodes with few incoming links. This seems to suggest that innovators are sought for advice and are often the only ones who can answer specific questions. In addition, innovators' ego-networks are more open to structural holes, probably leading to mediation advantages.

Innovators also write more, as they probably open new discussions when communicating innovation. Their language is more complex and articulated: They use longer words (higher language complexity) and have more words per sentence. Similarly, they also have a higher novelty score as a signal of new (less common) words being used, potentially related to new concepts or ideas introduced in the online discourse. Sentiment is, on average, positive for both for innovators and non-innovators, although the former had a lower average score (granted, this difference is not significant according to the Mann-Whitney's U test).

We extended our analysis by building logistic regression models to see which predictors, taken together, could better describe innovators' behavior. Table 5 presents the results.

|  | Model 1 | Model 2 | Model 3 | Model 4 | Model 5 | Model 6 |
|---|---|---|---|---|---|---|
| In-degree | -.00137* |  |  |  |  |  |
| Out-degree | .13134*** |  |  |  |  | .04660* |
| Weighted in-degree |  | -.00001 |  |  |  |  |
| Weighted out-degree |  | .01438*** |  |  |  |  |
| In-distinctiveness |  |  | -.00051** |  |  |  |
| Out-distinctiveness |  |  | .04424*** |  |  |  |
| Closeness |  |  |  | 7.16136*** |  | 6.37674*** |
| Betweenness |  |  |  | 4.62E-06 |  |  |
| Constraint |  |  |  | -1.86609*** |  | -1.53312** |
| Word Count |  |  |  |  | -.00006*** |  |
| Sentiment |  |  |  |  | -.58629 |  |
| Novelty |  |  |  |  | .60227*** |  |
| WPS |  |  |  |  | .01810*** | .01356** |
| Six-letters |  |  |  |  | .03731*** | .03438** |
| Constant | -5.70430*** | -5.43064*** | -5.53991*** | -7.60653*** | -8.19775*** | -9.05990*** |
| McFaddens' $R^2$ | 0.0475 | 0.0354 | 0.0532 | 0.3126 | 0.1717 | 0.3364 |

^ $p < .1$; * $p < .05$; ** $p < .01$; *** $p < .001$.

**Table 5**. Logistic regression models

We first evaluated the performance of our predictors in blocks, which helped to avoid collinearity issues. Indeed, putting the predictors of the first three models together would lead to high variance inflation factors (in this sample, we also find that in-distinctiveness is highly correlated with in-degree and weighted in- and out-degree). Model fit was evaluated according to McFadden's $R^2$, which reaches very good values in Models 4 and 6. We observe from the first three models that the highest $R^2$ values are obtained when using distinctiveness centrality in lieu of degree and weighted degree. Model 4 shows that closeness and constraint can play a significant role in describing the behavior of innovators. In Model 5, we look at language characteristics, which are all significant apart from sentiment. Model 6 is the best and most parsimonious model: It suggests that innovators are more central in the social network, having higher closeness and more brokerage power. Innovators also reach out to more colleagues (higher out-degree) and seem to write more complex posts (with more words of at least six letters and more words per sentence). Meanwhile, sentiment, novelty, and word count do not appear to play a significant role in distinguishing innovators from their peers.

## 5. Discussion and conclusions

### 5.1. How innovators differ from other employees

In order to find potential innovators within an organization, it is important to know how innovators behave—not only when pushed by the task to search for innovation, but also in informal social interactions. In this work, we traced the language and social behavior of "innovative employees" within the informal context of an online forum dedicated to sharing opinions, knowledge, and ideas. For this purpose, we analyzed the online communication of more than 11,000 employees working for a large multinational company. Our analysis was twofold: First, we conducted an in-depth study of the language used by innovators to profile them and see whether their communication style was different from those of other employees. Second, we tried to understand if innovators differ from their peers on several additional characteristics drawn from mapping their behaviors and social positioning in the network. We profiled these innovators by using a mix of tools from social network analysis, text mining and semantic analysis.

We found that innovators differ significantly from other employees on both the language and social behavior dimensions. Their interactions in the forum were mostly oriented toward exchanging ideas and opinions related to the company's growth and market opportunities: they tended to look outside the company boundaries rather than endorsing an inward orientation. They did not appear to be particularly interested in personal gain and company benefits; rather, they shared a vision of the potential benefits that the company's growth might generate for the community. Of course, this behavior increases business value and serves as a source of competitive advantage.

In terms of forum interactions, we found that innovators are substantially more active. They sent many more messages, wrote longer posts and used a more articulated and complex language. Thus, they signaled behaviors of open communication and inclusivity (Rogers et al., 2019). Their posts also showed higher novelty than non-innovators, insofar as they used more words departing from the common language.

From the results of Emotional Text Mining, we observed that innovators' messages are positive overall, but not as much as some non-innovators. They seem to engage in more objective and factual-based communication, have productivity-oriented communication, use words concerning business and marketing, and often report facts. Non-innovators, on the other hand, use the forum in a way that is more oriented toward socializing and building relationships or to discuss employees' benefits. Sometimes they show bursts of enthusiastic messages that feature little informative content. For example, some of their messages express appreciation for employee benefits (often including words such as "thank you" or "great!"), without adding much to the conversation.

Innovators are also more central in the social network and have more opportunities to bridge structural holes. Interestingly, they also have more distinctive outgoing social connections: In fact, they reach more peripheral employees whose questions would remain otherwise unanswered. This evidence suggests that innovators can provide valuable advice that cannot be searched elsewhere, which reflects prior evidence that innovators act as boundary-spanners and information-brokers within organizational boundaries (Fleming &

Waguespack, 2007; P. A. Gloor et al., 2020). The role they have within the organization might help to introduce pioneering initiatives, such as the adoption of breakthrough technologies (e.g., artificial intelligence technologies) (Pellikka & Ali-Vehmas, 2016) that are often not adopted because of technological, organizational, and environmental inhibitors (Enholm et al., 2021).

From a theoretical point of view, our work contributes to the innovation literature, which focuses on understanding innovators' characteristics with respect to less innovative people within an organization. Previous studies have argued that, when searching for innovative solutions, innovators tend to use a precise approach based on a set of activities and practices that make them more likely to discover relevant information, effectively elaborate it, and then connect dots in an original way to discover new solutions. While these studies have offered a valuable perspective to understanding where innovative and creative thinking come from, our study shifts the attention to how innovators reveal themselves in contexts that are not directly related to the search for innovation. The basic intuition is innovators can also manifest their characteristics in other social venues. In these settings, the interactions with a broad community may reveal behavioral patterns that are unique to creative minds. We specifically focused on language because it represents a social factor that guides individual and group cognition and, thus, represents the main means of social interaction.

### 5.2. Managerial implications

From a managerial point of view, this study offers interesting implications. Organizations are pushed to identify potential innovators among their employees in order to increase their innovativeness and competitiveness. Thus, identifying the attributes of innovators is critical for managerial decisions. Innovators, for instance, can help design and adopt innovative solutions, stimulate positive attitudes toward a new technology, and pave the way for identifying potential adopters of an innovative product. With our approach, we present measures and tools for identifying innovative employees simply by looking at how they interact and communicate with other organizational members.

By tracing innovators' profiles, organizations can better assess their ability to adopt a technology successfully, which encompasses all the organizational changes necessary for generating value from said technology (Collins et al., 2021). To continue with the previous example of AI, this technology has the potential to change innovation management practices and increase the effectiveness and efficiency of specific innovation tasks. However, AI-based innovation management asks for technical and organizational changes to cope with the challenges associated with its implementation. Organizations need to be structured with a pool of AI-innovators to succeed in this technological transition (Füller et al., 2022). Without innovators, the opportunity recognition associated with this technology would be limited. This could have severe implications for organizations' subsequent business value, competitive advantage, and corporate growth (Ahmed & Shepherd, 2010).

### 5.3. Limitations and avenues for future research

This study features some limitations that reveal future research avenues. A first limitation is that we analyzed the innovation profile of employees interacting in a rather informal online setting established by a specific multinational company. While we offer a methodology for analyzing the dominant themes that characterize innovators' language and social attitudes, the results cannot be generalized and may change from company to company. Therefore, we suggest replicating our methodology in other settings and recalibrating the profiling system depending on the specific context under investigation. Another limitation is that there might be the chance that some employees, who could have been qualified as innovators, did not use the online forum. However, we are rather confident that this situation is of limited concern for two reasons: First, the online forum was promoted by the company on several occasions. Second, employees with an intrinsic propensity to suggest ideas toward better practices, procedures, and products are comparatively less likely to disregard being engaged in public debates. In addition, because of the confidential nature of employees' information, we cannot link innovators' individual characteristics (such as education, gender, age, or tenure) to our metrics. Nonetheless, that information would have facilitated additional statistical tests (such as propensity score matching, as a way to handle group imbalance while evaluating mean differences via t-tests). Future research should investigate the extent to which other, more qualifying individual features relate to online social behavior and language style. Lastly, we had no information on post visualizations/reads or likes, as well as on private messages that employees could exchange on the forum. Future research could consider these metrics as a proxy for post and employee popularity, in addition to the factors considered here (the number of public responses received by each employee and the number of different respondents). Indeed, looking at private messages and likes could provide additional data to extend the representation of the online social network.

Ultimately, we believe that our analytical approach can be useful for profiling innovators within a company context. The presented measures and tools can be especially valuable for identifying innovators who are unknown in the company or when a generalized expert assessment is not a viable option (e.g., due to time or cost constraints). We hope this study will spur future research that considers larger datasets and new settings in order to enhance our understanding of innovators' language and social behavior.

**Appendix**

**Table 1A.** Literature review's summary

| Macro-area | Sub-investigation areas | References |
|---|---|---|
| *The communication style of innovators* | | |
| Communication dynamics and organizational performance | i. Communication channels, R&D networks/organizations, and technological performance | Allen et al., 2016; Fosfuri & Tribó, 2008; Sousa & Rocha, 2019 |
| | ii. Communication networks, and organizational change | Harandi and Abdolvand, 2018; Raz & Gloor, 2007 |
| | iii. Communication, leadership, and organizational change | Alonderiene et al., 2020; Hsing-Er Lin & McDonough, 2011; Schaufeli, 2013; Xu & Cooper Thomas, 2011 |
| | iv. Communication environment, language, and engagement | Morrison, 2014; Johnston et al. 2019; Rees et al., 2013; Reissner & Pagan, 2013; Van Dyne, Ang, and Botero, 2003; |
| Language style and innovation strategies | i. Language style, network dynamics, and innovation | Dyer et al., 2011; Gloor et al., 2020; Greco et al. 2021; Ray et al., 1996; Weber and Grauer, 2019 |
| | ii. Language-based audience adaptation and information processing strategies, and innovation dynamics | Bazarova et al., 2013; Dyer et al., 2011; Gil-Lopez et al., 2018 |
| *The social behavior of innovators* | | |
| Organizational social interactions, knowledge creation/diffusion and innovation | i. Social network structure and knowledge diffusion | Carpenter et al., (2012); Cross & Prusak, 2002; Cullen et al., 2018; Dobrow et al. 2012; Ibarra et al., 1993; Lee et al., 2011; Phelps et al. 2012; Shane & Cable, 2002; Tsai, 2001; Zasa et al., 2022 |
| | ii. Social network structure and innovation | Cangialosi et al., 2021; Ibarra et al., 1993; Deeds et al., 2004; Elfring & Hulsink, 2007; Gulati & Higgins, 2003; Muller and Peres, 2019; Nedkovski and Guerci, 2021; Shane & Cable, 2002; Valente, 1996; West & Richter, 2008 |

| Organizational social network, personality traits and cognitions | i. | Network embeddedness, behaviors and personality | Battistoni & Fronzetti Colladon, 2014; Fang et al., 2015; Fronzetti Colladon et al., 2021; Kilduff et al., 2008; Landis, 2016; Mehra et al., 2001; Tasselli et al., 2015; Schulte et al., 2012; Yan et al., 2020 |
|---|---|---|---|
| Social network structure and boundary conditions | i. | Network structural characteristics and boundary conditions | Bellavitis et al., 2017; Gulati & Higgins, 2003; Kilduff & Brass, 2010; Shipilov & Li, 2008 |

**References**

Ahmed, P., & Shepherd, C. (2010). *Innovation Management: Context, Strategies, Systems and Processes*. Pearson Education Limited.

Ahuja, M. K., Galletta, D. F., & Carley, K. M. (2003). Individual Centrality and Performance in Virtual R&D Groups: An Empirical Study. *Management Science*, *49*(1), 21–38. https://doi.org/10.1287/mnsc.49.1.21.12756

Allen, T. J., Gloor, P., Fronzetti Colladon, A., Woerner, S. L., & Raz, O. (2016). The power of reciprocal knowledge sharing relationships for startup success. *Journal of Small Business and Enterprise Development*, *23*(3), 636–651. https://doi.org/10.1108/JSBED-08-2015-0110

Alonderiene, R., Muller, R., Pilkiene, M., Simkonis, S., & Chmieliauskas, A. (2020). Transitions in Balanced Leadership in Projects: The Case of Horizontal Leaders. *IEEE Transactions on Engineering Management*, 1–13. https://doi.org/10.1109/TEM.2020.3041609

Amabile, T. (1988). A Model of Creativity and Innovation in Organizations. *Research in Organizational Behavior*, *10*, 123–167.

Battistoni, E., & Fronzetti Colladon, A. (2014). Personality correlates of key roles in informal advice networks. *Learning and Individual Differences*, *34*, 63–69. https://doi.org/10.1016/j.lindif.2014.05.007

Bazarova, N. N., Taft, J. G., Choi, Y. H., & Cosley, D. (2013). Managing Impressions and Relationships on Facebook. *Journal of Language and Social Psychology*, *32*(2), 121–141. https://doi.org/10.1177/0261927X12456384

Bellavitis, C., Filatotchev, I., & Souitaris, V. (2017). The Impact of Investment Networks on
36